\documentclass[conference]{IEEEtran}
\IEEEoverridecommandlockouts
\usepackage{cite}
\usepackage{amsmath,amssymb,amsfonts}
\usepackage{algorithmic}
\usepackage{graphicx}
\usepackage{textcomp}
\usepackage{xcolor}

\usepackage{diagbox}
\usepackage[acronym]{glossaries}
\usepackage{siunitx}
\usepackage{amsthm}
\usepackage{svg}
\usepackage{multirow}
\usepackage{tikz}
\usepackage{pgfplots}
\usepackage{subfig}
\usepackage{array}
\usepackage{xcolor}
\usepackage{booktabs}
\usepackage{ragged2e}
\usepackage{comment}
\PassOptionsToPackage{hyphens}{url}\usepackage{hyperref}

\usepackage{fancyhdr}

\usepackage[a4paper, total={184mm,239mm}]{geometry}

\def\BibTeX{{\rm B\kern-.05em{\sc i\kern-.025em b}\kern-.08em
    T\kern-.1667em\lower.7ex\hbox{E}\kern-.125emX}}

\fancypagestyle{FirstPage}{
\cfoot{\footnotesize © 2024 IEEE.  Personal use of this material is permitted.  Permission from IEEE must be obtained for all other uses, in any current or future media, including reprinting/republishing this material for advertising or promotional purposes, creating new collective works, for resale or redistribution to servers or lists, or reuse of any copyrighted component of this work in other works.} 
}

\begin{document}

\title{12 mJ per Class On-Device Online \\ Few-Shot Class-Incremental Learning}

\author{
Yoga Esa Wibowo{$\ddagger$}, Cristian Cioflan*{$\dagger$}, Thorir Mar Ingolfsson{$\dagger$}, Michael Hersche{$\dagger$}{$\mathparagraph$}, \\ 
Leo Zhao{$\ddagger$}, Abbas Rahimi{$\mathparagraph$}, Luca Benini{$\dagger$}{$\mathsection$} \\
\IEEEauthorblockA{\textit{
\textsuperscript{$\ddagger$}D-ITET, ETH Zurich; 
\textsuperscript{$\dagger$}Integrated Systems Laboratory, ETH Zurich; \textsuperscript{$\mathparagraph$}IBM Research-Zurich; \textsuperscript{$\mathsection$}DEI, University of Bologna}}
\IEEEauthorblockA{\small \{ywibowo, lezhao\}@ethz.ch, \{cioflanc, thoriri, lbenini\}@iis.ee.ethz.ch, \{her, abr\}@zurich.ibm.com}

\thanks{*~Corresponding author}
}

\maketitle

\newacronym[plural=FLLs,firstplural=Frequency Locked Loops (FLLs)]{fll}{FLL}{Frequency Locked Loop}
\newacronym{inq}{INQ}{Incremental Network Quantization}
\newacronym{tqt}{TQT}{Training Quantization Thresholds}
\newacronym{ste}{STE}{Straight-Through-Estimator}
\newacronym{bn}{BN}{Batch Normalization}
\newacronym{dma}{DMA}{Direct Memory Access}
\newacronym{simd}{SIMD}{Single Instruction Multiple Data}
\newacronym{lr}{LR}{Learning Rate}
\newacronym[plural=PTUs, firstplural={Pan-Tilt Units}]{ptu}{PTU}{Pan-Tilt Unit}
\newacronym{mdf}{MDF}{Medium-density fibreboard}
\newacronym{soa}{SoA}{State of the Art}
\newacronym{lorawan}{LoRaWAN}{Long Range Wide Area Network}
\newacronym{lora}{LoRa}{Long Range}
\newacronym{dram}{DRAM}{Dynamic Random Access Memory}
\newacronym{fpu}{FPU}{Floating Point Unit}
\newacronym[plural=SCMs, firstplural={Standard Cell Memories (SCMs)}]{scm}{SCM}{Standard Cell Memory}

\newacronym[plural=DVS, firstplural={Dynamic Vision Sensors (DVS)}]{dvs}{DVS}{Dynamic Vision Sensor}
\newacronym[plural=FPGAs, firstplural={Field Programmable Gate Arrays (FPGAs)}]{fpga}{FPGA}{Field Programmable Gate Array}
\newacronym{tpu}{TPU}{Tensor Processing Unit}
\newacronym[plural=WANs, firstplural={Wide Area Networks (WANs)}]{wan}{WAN}{Wide Area Network}
\newacronym[plural=WSNs, firstplural={Wireless Sensor Networks (WSNs)}]{wsn}{WSN}{Wireless Sensor Network}
\newacronym{dl}{DL}{Deep Learning}

\newacronym{fbk}{FBK}{Fondazione Bruno Kessler}
\newacronym{FGSM}{FBK}{Fast Gradient Sign Method}
\newacronym{date}{DATE}{Design Automation and Test in Europe}
\newacronym{iimtc}{I2MTC}{The International Instrumentation \& Measurement Technology Conference}
\newacronym{ini}{INI}{Institute of Neuroinformatics}
\newacronym[plural=LUTs, firstplural={Lookup Tables (LUTs)}]{lut}{LUT}{Lookup Table}

\newacronym{lpwan}{LPWAN}{Low-Power Wide Area Network}
\newacronym{nbiot}{NB-IoT}{Narrow Band Internet-of-Things}

\newacronym{saer}{SAER}{Synchronous Address-Event Representation}
\newacronym{fps}{FPS}{Frames Per Second}
\newacronym{vcd}{VCD}{Value-Change Dump}
\newacronym{spi}{SPI}{Serial Peripheral Interface}
\newacronym{cpi}{CPI}{Camera Parallel Interface}
\newacronym{fifo}{FIFO}{First-In First-Out Queue}
\newacronym{ble}{BLE}{Bluetooth Low-Energy}
\newacronym{wifi}{Wi-FI}{Wireless Fidelityy}
\newacronym{uart}{UART}{Universal Asynchronous Receiver-Transmitter}
\newacronym{sta}{STA}{Static Timing Analysis}
\newacronym{ptz}{PTZ}{Pan-Tilt Unit}
\newacronym[plural=GPIOs, firstplural={General Purpose Inupt Outputs (GPIOs)}]{gpio}{GPIO}{General Purpose Input Output}
\newacronym[plural=LDOs, firstplural={Low Dropout Regulators (LDOs)}]{ldo}{LDO}{Low Dropout Regulator}

\newacronym{CV}{CV}{Computer Vision}
\newacronym{EoT}{EoT}{Expectation over Transformation}
\newacronym{RPN}{RPN}{Region Proposal Network}
\newacronym{TV}{TV}{Total Variation}
\newacronym{NPS}{NPS}{Non-Printability Score}
\newacronym{STN}{STN}{Spatial Transformer Network}
\newacronym{MTCNN}{MTCNN}{Multi-Task Convolutional Neural Network}
\newacronym{YOLO}{YOLO}{You Only Look Once}
\newacronym{SSD}{SSD}{Single Shot Detector}
\newacronym{SOTA}{SOTA}{State of the Art}
\newacronym{NMS}{NMS}{Non-Maximum Suppression}
\newacronym{ic}{IC}{Integrated Circuit}
\newacronym{rf}{RF}{Radio Frequency}
\newacronym{tcxo}{TCXO}{Temperature Controlled Crystal Oscillator}
\newacronym{jtag}{JTAG}{Joint Test Action Group industry standard}
\newacronym{swd}{SWD}{Serial Wire Debug}
\newacronym{sdio}{SDIO}{Serial Data Input Output}

\newacronym[plural=PCBs, firstplural={Printed Circuit Boards (PCB)}]{pcb}{PCB}{Printed Circuit Board}
\newacronym[plural=ASICs, firstplural={Application Specific Integrated Circuits}]{asic}{ASIC}{Application Specific Integrated Circuit}

\newacronym{ml}{ML}{Machine Learning}
\newacronym{ai}{AI}{Artificial Intelligence}
\newacronym{iot}{IoT}{Internet of Things}
\newacronym{fft}{FFT}{Fast Fourier Transform}
\newacronym[plural=OCUs, firstplural={Output Channel Compute Units (OCUs)}]{ocu}{OCU}{Output Channel Compute Unit}
\newacronym{alu}{ALU}{Arithmetic Logic Unit}
\newacronym{mac}{MAC}{Multiply-Accumulate}
\newacronym{soc}{SoC}{System-on-Chip}

\newacronym{PGD}{PGD}{Projected Gradient Descend}
\newacronym{CW}{CW}{Carlini-Wagner}
\newacronym{OD}{OD}{Object Detection}

\newacronym{rrf}{RRF}{RADAR Repetition Frequency}
\newacronym{nlp}{NLP}{Natural Language Processing}
\newacronym{qam}{QAM}{Quadrature Amplitude Modulation}
\newacronym{rri}{RRI}{RADAR Repetition Interval}
\newacronym{radar}{RADAR}{Radio Detection and Ranging}
\newacronym{loocv}{LOOCV}{Leave-one-out cross validation}
\newacronym{raw}{RAW}{Read-After-Write}
\newacronym[plural=ISAs, firstplural={Instruction Set Architectures (ISAs)}]{isa}{ISA}{Instruction Set Architecture}

\newacronym{os}{OS}{Operating System}
\newacronym{bsp}{BSP}{Board Support Package}
\newacronym{ttn}{TTN}{The Things Network}
\newacronym{wip}{WIP}{Work in Progress}
\newacronym{json}{JSON}{JavaScript Object Notation}
\newacronym{qat}{QAT}{Quantization-Aware Training}

\newacronym{cls}{CLS}{Classification Error}
\newacronym{loc}{LOC}{Localization Error}
\newacronym{bkgd}{BKGD}{Background Error}

\newacronym{dsp}{DSP}{Digital Signal Processing}
\newacronym{mcu}{MCU}{Microcontroller Unit}

\newacronym{gsc}{GSC}{Google Speech Commands}
\newacronym{mswc}{MSWC}{Multilingual Spoken Words Corpus}
\newacronym{demand}{DEMAND}{Diverse Environments Multichannel Acoustic Noise Database}
\newacronym{coco}{COCO}{Common Objects in Context}

\newacronym{asr}{ASR}{Automated Speech Recognition}
\newacronym{kws}{KWS}{Keyword Spotting}
\newacronym{nl-kws}{NL-KWS}{Noiseless Keyword Spotting}
\newacronym{na-kws}{NA-KWS}{Noise-Aware Keyword Spotting}
\newacronym{odda}{ODDA}{On-Device Domain Adaptation}

\newacronym{snr}{SNR}{Signal-to-Noise Ratio}
\newacronym{roc}{ROC}{Receiver Operating Characteristic}
\newacronym{frr}{FRR}{False Rejection Rate}
\newacronym{eer}{EER}{Equal Error Rate}
\newacronym{ce}{CE}{Cross-Entropy}

\newacronym[plural=BNNs, firstplural={Binary Neural Networks (BNNs)}]{bnn}{BNN}{Binary Neural Network}
\newacronym[plural=NNs, firstplural={Neural Networks}]{nn}{NN}{Neural Network (NNs)}
\newacronym[plural=SNNs, firstplural={Spiking Neural Networks (SNNs)}]{snn}{SNN}{Spiking Neural Network}
\newacronym[plural=DNNs, firstplural={Deep Neural Networks (DNNs)}]{dnn}{DNN}{Deep Neural Network}
\newacronym[plural=TCNs,firstplural=Temporal Convolutional Networks]{tcn}{TCN}{Temporal Convolutional Network}
\newacronym[plural=CNNs,firstplural=Convolutional Neural Networks (CNNs)]{cnn}{CNN}{Convolutional Neural Network}
\newacronym[plural=TNNs,firstplural=Ternarized Neural Networks]{tnn}{TNN}{Ternarized Neural Network}
\newacronym{ann}{ANN}{Artificial Neural Networks}
\newacronym{ds-cnn}{DS-CNN}{Depthwise Separable Convolutional Neural Network}
\newacronym{rnn}{RNN}{Recurrent Neural Network}
\newacronym{gcn}{GCN}{Graph Convolutional Network}
\newacronym{mhsa}{MHSA}{Multi-Head Self Attention}
\newacronym{crnn}{CRNN}{Convolutional Recurrent Neural Network}
\newacronym{mhsa}{MHSA}{Multi-Head Self-attention}
\newacronym{clca}{CLCA}{Convolutional Linear Cross-Attention}
\newacronym{cvat}{CVAT}{Computer Vision Annotation Tool}

\newacronym{mfcc}{MFCC}{Mel-Frequency Cepstral Coefficient}
\newacronym{bf}{BF}{Beamforming}
\newacronym{anc}{ANC}{Active Noise Cancellation}
\newacronym{agc}{AGC}{Automatic Gain Control}
\newacronym{se}{SE}{Speech Enhancement}
\newacronym{mct}{MCT}{Multi-Condition Training}
\newacronym{mcta}{MCTA}{Multi-Condition Training \& Adaptation}
\newacronym{pcen}{PCEN}{Per-Channel Energy Normalization}
\newacronym{sf}{SF}{Sensor Fusion}

\newacronym{mac}{MAC}{Multiply-Accumulate}
\newacronym{flop}{FLOP}{Floating-Point Operation}
\newacronym{fp}{FP}{Floating-Point}

\newacronym{fscil}{FSCIL}{Few-Shot Class-Incremental Learning}
\newacronym{ofscil}{O-FSCIL}{Online Few-Shot Class-Incremental Learning}
\newacronym{ncfscil}{NC-FSCIL}{Neural Collapse Few-Shot Class-Incremental Learning}
\newacronym{cfscil}{C-FSCIL}{Constrained Few-Shot Class-Incremental Learning}
\newacronym{fsl}{FSL}{Few-Shot Learning}
\newacronym{cil}{CIL}{Class-Incremental Learning}
\newacronym{dil}{DIL}{Domain-Incremental Learning}
\newacronym{savc}{SAVC}{Semantic-Aware Virtual Contrastive}

\newacronym{am}{AM}{Activation Memory}
\newacronym{em}{EM}{Explicit Memory}
\newacronym{fcr}{FCR}{Fully Connected Reductor}
\newacronym{fcc}{FCC}{Fully Connected Classifier}

\begin{abstract}

\gls{fscil} enables machine learning systems to expand their inference capabilities to new classes using only a few labeled examples, without forgetting the previously learned classes. Classical backpropagation-based learning and its variants are often unsuitable for battery-powered, memory-constrained systems at the extreme edge. In this work, we introduce~\gls{ofscil}, based on a lightweight model consisting of a pretrained and metalearned feature extractor and an expandable explicit memory storing the class prototypes. The architecture is pretrained with a novel feature orthogonality regularization and metalearned with a multi-margin loss. For learning a new class, our approach extends the explicit memory with novel class prototypes, while the remaining architecture is kept frozen. This allows learning previously unseen classes based on only a few examples with one single pass (hence online).~\gls{ofscil} obtains an average accuracy of 68.62\% on the~\gls{fscil} CIFAR100 benchmark, achieving state-of-the-art results. Tailored for ultra-low-power platforms, we implement~\gls{ofscil} on the 60\,mW GAP9 microcontroller, demonstrating online learning capabilities within just 12\,mJ per new class.

\end{abstract}


\begin{IEEEkeywords}
Continual learning, on-device learning, deep neural networks, microcontrollers.
\end{IEEEkeywords}

\thispagestyle{FirstPage}
\section{Introduction}
\label{sec:introduction}

Classical~\gls{ml} solutions often use large datasets to train a highly complex yet fixed model, which cannot adapt to the needs and requirements of the end user. 
Such systems are nonetheless exposed to dynamic, ever-changing environments; thus, adaptability is a crucial requirement for an intelligent system. 
Moreover, while in a server-based, offline learning paradigm, curated and labeled data is widely available, that is seldom the case when a pretrained model must adapt to a particular user.
~\glsfirst{fscil} evaluates models on two aforementioned problems, requiring a previously trained model to expand its classification domain while providing very few labeled training samples. 
The key challenge is to prevent catastrophic forgetting (i.e., forgetting prior knowledge during knowledge expansion) and avoid overfitting to the few novel samples. 
These goals can be achieved by balancing stability and plasticity~\cite{tao2020fscil}. 
Several~\gls{fscil} solutions have tackled sample-scarce class learning over several incremental sessions~\cite{zhang2021few,zhu2021self} in a transfer learning fashion, where a model is pretrained to classify a set of predefined base classes, followed by freezing its backbone and only training its classification head to learn novel classes.

Although such methods achieve remarkable results on previously unseen classes, they rely on the computationally expensive backpropagation algorithm.
Alternative solutions, such as~\gls{ncfscil}~\cite{yang2023ncfscil},~\gls{savc} model~\cite{song2023savcfscil}, or~\gls{cfscil}~\cite{hersche2022constrained,karunaratne2022memory}, use large ResNet backbones~\cite{he2016deep}. 
However, adapting to new requirements and learning novel classes should happen directly on the user device, thus addressing privacy and security concerns~\cite{kumar2019iot}, while also avoiding energy-hungry data streaming to the cloud. 
Therefore, when employed at the extreme edge,~\gls{fscil} algorithms should operate under the TinyML constraints~\cite{banbury2020benchmarking}; hence, suitably small and computationally affordable backbones are required.

On-device training frameworks targeting~\glspl{mcu} have been proposed~\cite{ren2021tinyol, nadalini2022pulptrainlib, lin2022ondevice}, which encouraged the development of domain adaptation~\cite{cioflan2022towards} and class continual learning~\cite{ravaglia2021tinyml} strategies tailored for~\glspl{mcu}. 
However, such iterative learning approaches require storing training samples during the update process.
Our proposed online training methodology addresses the storage constraints by enabling novel class learning with a single pass over the available samples.
This also decreases the learning time and, thus, the energy consumption of our system.
Moreover, as opposed to~\gls{fscil} strategies deployed on~\glspl{tpu}~\cite{lungu2020siamese}, ~\glspl{fpga}~\cite{chen2021eile}, neuromorphic chips~\cite{hajizada2022interactive}, or in-memory processors~\cite{karunaratne2022memory}, our methodology is suitable for off-the-shelf, widely available~\glspl{mcu}, with a conservative power profile.

This work introduces~\glsfirst{ofscil}\footnote{The code will be open-sourced at: https://github.com/pulp-platform/fscil}, a lightweight~\gls{fscil} learning methodology aimed at resource-constrained edge devices. 
Using orthogonal regularization and data augmentation strategies such as Mixup~\cite{zhang2017mixup} and Cutmix~\cite{yun2019cutmix}, we pretrain the MobileNetV2~\cite{sandler2018mobilenetv2} backbone, which requires only 2.5 million parameters and 149.2~million \gls{mac} operations. 
During the server-side metalearning phase, we employ a multi-margin loss to prevent overfitting to the few labeled samples. 
We further demonstrate the on-device learning capabilities by deploying and evaluating~\gls{ofscil} on ultra-low-power, commercially available GAP9~\cite{gap9datasheet}.

The main contributions of the paper are as follows:
\begin{itemize}
    \item We introduce~\gls{ofscil}, a class-incremental learning method achieving a new state-of-the-art average accuracy of 68.62\% on the~\gls{fscil} CIFAR100~\cite{krizhevsky2009cifar100} benchmark.
    \item For each novel class, our pretrained and metalearned backbone generates orthogonal feature vectors quantized to 3-bit integers, which are stored in the~\gls{em}, resulting in memory requirements of only \SI{9.6}{\kilo \byte} for 100 classes.
    \item We demonstrate few-shot online (i.e., single-pass) learning capabilities on a \SI{50}{\m \watt}, requiring as little as \SI{12}{\m \joule} to learn a new class.
\end{itemize}

\section{Related Work}
\label{sec:relatedwork}

\subsection{Few-shot Class-incremental Learning}
\label{ssc:fewshot}

Different approaches have been proposed to tackle the~\gls{fscil} scenario, in which models presented with out-of-distribution data learn new classes from few labeled data.
TOPIC~\cite{tao2020fscil} introduces the~\gls{fscil} problem and employs a neural gas technique to maintain the topology in the embedding space.
By updating the backbone of the neural network~\cite{tao2020fscil, dong2021few, zhao2021mgsvf}, one can successfully learn to recognize previously unseen classes.
To learn new classes and avoid catastrophic forgetting without maintaining a large reservoir memory, we freeze the backbone and only store class prototypes in the explicit memory.

To reduce the costs of retraining the backbone, other works~\cite{zhang2021few,zhu2021self} proposed to freeze the backbone and operate on the classifier and its additional components.
Works such as~\cite{zhou2022limit, zhang2021few, zhu2021self} rely on an episodic memory, where class-specific information is stored and compared with the query image during inference.
Zhou et al.~\cite{zhou2022limit} introduce a forward compatibility methodology, where a provident model minimizes the negative compression effects novel classes have on the embedding space.
ALICE~\cite{peng2022alice} proposes the angular penalty loss to achieve compact clustering and feature diversity, achieving better generalization for unseen classes.
Conversely, \gls{savc}~\cite{song2023savcfscil} enhance the cluster separation with virtual classes created through predefined transformations, which diversify the semantic information.
However, these methods do not generate meaningful feature representations in space between class clusters, thus impeding clustering in that vacant space. 

NC-FSCIL~\cite{yang2023ncfscil} tackles the clustering problem by creating a placeholder for all class prototypes with fixed, predetermined vectors, thus addressing the prototype readjustment issue when adding a new class cluster.
As opposed to these works, our methodology improves features' representations and expressiveness already during the pretraining and metalearning stages, through feature interpolation, as well as orthogonality and multi-margin loss-based network update.
This allows us to achieve higher accuracy levels with inexpensive few-shot class-incremental learning, suitable for on-device deployment.

\subsection{On-device continual learning}
\label{ssc:ondevice}

While deployment frameworks enabling~\gls{dnn} inference are already well established~\cite{burrello2020dory}, on-device training frameworks~\cite{ren2021tinyol,lin2022ondevice} are yet to generalize to multple~\gls{dl} topologies and generally target single-platform families.
Notably, Nadalini et al.~\cite{nadalini2022pulptrainlib} propose a framework aimed at both single- and multi-core platforms, accounting for the particular memory hierarchy of each target device.
To improve accessibility, in~\cite{pellegrini2021continual} the authors introduce a continual learning framework for smartphones.
Conversely, Chen et al.~\cite{chen2021eile} introduce a reconfigurable array architecture to accelerate backpropagation through uniform memory access patterns, with \SI{410}{\m \watt} power consumption on~\gls{fpga}.
Nonetheless, such works address backpropagation-based training, which is generally unfit for extreme edge devices due to memory and computational requirements. Furthermore, backpropagation-based fine-tuning requires large amounts of labeled data, often unavailable in real-world settings.

In order to solve the extreme edge learning challenges, several on-device (few-shot) class-incremental learning implementations have been proposed.
Hacene et al.~\cite{hacene2017incremental} introduce a feature extractor whose features are compared with class anchor points, their architecture implemented in a \SI{22}{\watt}~\gls{fpga}. 
Lungu et al.~\cite{lungu2020siamese} use Siamese Networks to compute the similarity measure between a query and class prototypes, with their~\gls{fpga} implementation enabling class learning within \SI{35}{\m \s}.
Neuromorphic chips have also been employed for continual~\cite{hajizada2022interactive} and few-shot~\cite{stewart2020online} on-device learning, with energy requirements of \SI{167}{\m \joule} per class in a data-scarce context.
Instead, we propose a methodology enabling real-time inference and learning, with energy requirements as low as \SI{12}{\m \joule}.

\section{\gls{fscil} task description}
\label{ssc:fsciltask}

In~\gls{fscil}, a learner is progressively presented with new classes over a number of training sessions. 
The data stream is denoted as $D={\{D^{t}\}}_{t=0}^{T}$, where $t$ indicates the session index.
The class labels $y_n^t\in{\mathcal{C}}^{t}$ does not intersect across sessions, i.e., $\forall i\neq j, \mathcal{C}^i \cap \mathcal{C}^j = \emptyset$.
During the base session, or session 0, the model is pretrained and metalearned to gain representation knowledge of the input sequence.
Subsequently, in the online stage, the incremental sessions introduce $N$ new classes, each with $S$ samples per class, referred to as N-way, S-shot~\gls{fscil}.
The model is evaluated on samples from all previous classes $\widetilde{C^t}:=C^{0}\cup C^{1}\cdot\cdot\cdot\cup C^{t}$, ensuring that new classes are learned without forgetting previous ones.

\section{Online Few-shot Class-incremental Learning}
\label{sec:software}

\begin{figure*}[t]
    \centering
    \subfloat[Classifying a query in inference mode by comparing its features with class prototypes stored in EM.]{%
        \includegraphics[width=0.3\textwidth]{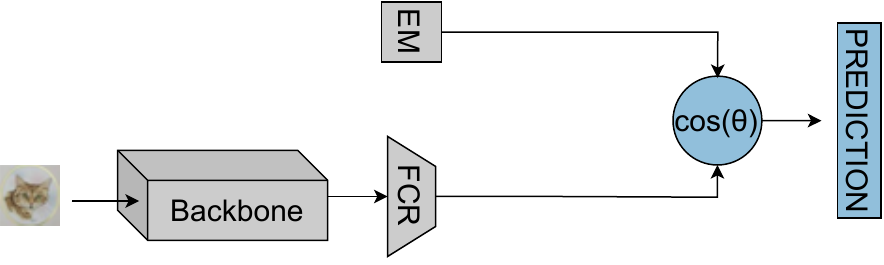}%
        \label{fig:architecture_inference}%
    }\hfill
    \subfloat[O-FSCIL update of EM with the class average of FCR-generated features. The backbone and the FCR are frozen. ]{%
        \includegraphics[width=0.3\textwidth]{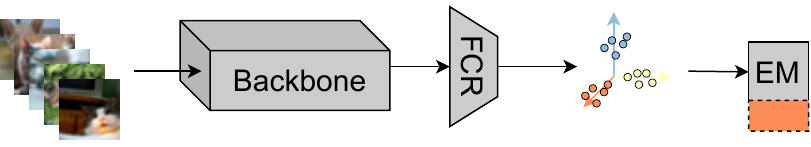}%
        \label{fig:architecture_incremental}%
    }\hfill
    \subfloat[Server-side metalearning the model on the base session. The loss is used to update the backbone and the FCR.]{%
        \includegraphics[width=0.3\textwidth]{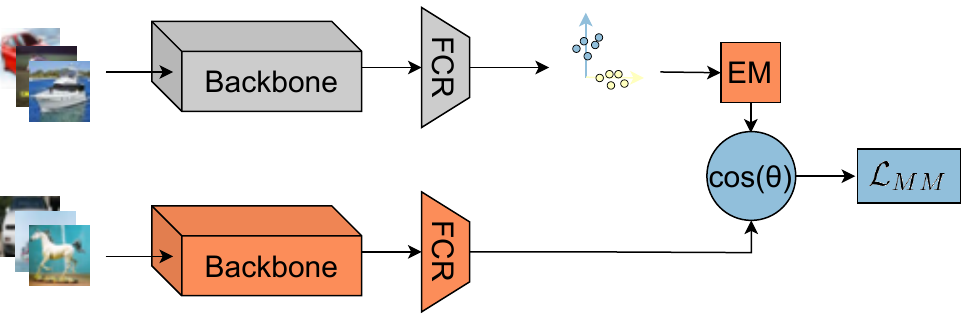}%
        \label{fig:architecture_metalearning}%
    }
    \caption{Inference (a), on-device learning a new class (b), and server-side metalearning (c) modes of O-FSCIL. Modules colored in orange are updated, while grey ones are frozen. During pretraining, we replaced the prototype computation and EM update from (b) with an FCR-like FCC classifier, with all three sections jointly trained.}
    \label{fig:architecture}

\vspace{-0.5cm}
    
\end{figure*}

This section presents the first main contribution of the paper: we introduce~\gls{ofscil}, which is comprised of a backbone, a~\gls{fcr}, and an~\glsfirst{em}, shown in  Fig.~\ref{fig:architecture}. 
The backbone ($f(\cdot)$) maps an input image ($x$) to an intermediate representation ($\theta_a \in \mathbb{R}^{d_a}$), and the~\gls{fcr} projects the representation to a lower-dimensional feature ($\theta_p \in \mathbb{R}^{d_p}$ where $d_p < d_a$). 
During inference, the feature vector ($\theta_p$) is compared against all prototypes stored in the~\gls{em}; the prototype with the highest cosine similarity indicates the final prediction (see Fig.~\ref{fig:architecture_inference}).

When learning a new class $i$, the prototype ($\overline\theta_p,i$) is added to the~\gls{em} through averaging all $\theta_p$ feature of class $i$ samples, while the backbone and~\gls{fcr} remains frozen.
This allows for online updates by passing labeled images through the model only once, without requiring expensive iterative (batched) gradient updates. 
This online learning capability is acquired via pretaining and metalearning, which are performed before deploying the model using data from the base session.

\subsection{Light-weight backbone}
\label{ssc:architecture}

To enable~\gls{fscil} at the extreme edge, we replace the ResNet-12 backbone~\cite{hersche2022constrained,song2023savcfscil} with a lightweight MobileNetV2~\cite{sandler2018mobilenetv2}.
To handle low-resolution inputs (i.e., 32$\times$32 in CIFAR100~\cite{krizhevsky2009cifar100}), we reduce the stride across the convolutional blocks of the seven inverted residual blocks, obtaining three models with different complexity.
The proposed networks and their hardware costs are presented in Table~\ref{tab:3_modified_mobilenetv2}, together with the dimensions of the~\gls{fcr} features ($d_a$) and prototypical features ($d_p$).

\subsection{Pretraining}
\label{ssc:pretraining}

The pretraining guides the backbone and~\gls{fcr} towards meaningful representations by solving a supervised classification problem on the base session (with $|C^0|$ classes). 
To this end, we modify the topology shown in Fig.~\ref{fig:architecture_incremental} by replacing the~\gls{em} with a~\gls{fcc}, which computes the class probability of an input image. 
The backbone, the~\gls{fcr}, and the~\gls{fcc} are mutually trained to minimize the~\gls{ce} loss.
To improve the accuracy of our system, we implement data augmentation methods enabling feature interpolation and a novel feature orthogonality regularization. 

\begin{table}[t]
\caption{Proposed backbones. Here we represent the convolutional stride per inverted residual block in MobileNetV2, as well as the dimensionality of the~\gls{fcr} features $d_a$  and prototypical features $d_p$.}
\label{tab:3_modified_mobilenetv2}
\begin{tabular}{c|ccc|c}
\multicolumn{1}{c|}{{\textbf{}}} & \multicolumn{3}{c|}{\textbf{MobileNetV2}} & \textbf{Resnet12} \\ 
\multicolumn{1}{c|}{}  & \textbf{}  & \textbf{\_x2}      & \textbf{\_x4}      & \\ \hline
CNN stride & 1,2,2,2,1,2,1 & 1,2,2,2,1,1,1 & 1,2,2,1,1,1,1& - \\ \hline
$d_a$ & 1280 & 1280 & 1280 & 640 \\ \hline
$d_p$ & 256 & 256 & 256 & 512 \\ \hline
Params. [M] & $2.5$ & $2.5$ & $2.5$ & $12.9$ \\ \hline
MACs [M] & $25.9$ & $45.4$ & $149.2$ & $525.3$ \\ \hline
\end{tabular}

\vspace{-0.3cm}

\end{table}

\textbf{Feature interpolation}: Apart from traditional data augmentation techniques (i.e., blur, horizontal flip, crop, and resize), we use inter-class feature interpolation using Mixup \cite{zhang2017mixup} and CutMix \cite{yun2019cutmix}.
Thus, instead of generating multiple inputs from a single image, we combine two input images and create intermediate class labels.
The two methods are employed exclusively, with a probability of 0.4.

\textbf{Feature orthogonality}: 
The pretraining of the architecture with the~\gls{ce} loss can reduce the representationality of the prototype features.
The~\gls{fcc} layer reduces the dimension of $\theta_p$ from $\mathbb{R}^{d_p}$ to $\mathbb{R}^{|C^0|}$, where $|C^0|$ is the number of the base classes and $d_p > |C^0|$.
The classification boundary of~\gls{fcc} only lives in the smaller base class hyperplane with dimension equal to $|C^0|$, ignoring perpendicular features.
This impedes the generation of new feature clusters on orthogonal planes, hindering the model's ability to learn new classes.

We propose feature orthogonality regularization to address the dimensionality reduction problem.
Instead of applying weight matrix orthogonal regularization~\cite{ranasinghe2021orthogonal1}, we orthogonalize the feature vectors and study its generalization capability to novel classes in the~\gls{fscil} scenario.
Equation \ref{eq:pretrain_ortho_reg} shows the formulation of our orthogonality regularization:

\begin{equation}\label{eq:pretrain_ortho_reg}
    \mathcal{L}_{ortho} = (\theta_{pb}^t \times \theta_{pb} - I_{d_f})^2,
\end{equation}
where $\theta_{pb}$ is a batch of $\theta_p$ bundled as matrix with dimension $\mathbb{R}^{B\times d_p}$, and $B$ is the batch size. 
The proposed pretraining loss includes the classification~\gls{ce} loss and the orthogonality loss, weighted by the $\lambda_{ortho}$ regularization strength:

\begin{equation}\label{eq:loss_pretrain}
    \mathcal{L}_{pre} = \mathcal{L}_{ce} + \lambda_{ortho}\cdot\mathcal{L}_{ortho},
\end{equation}

\subsection{Metalearning}
\label{ssc:metalearning}

After pretraining, we perform offline metalearning to enhance feature clustering by emulating the learning and inference processes on the base session~\cite{chi2022metafscil}, as shown in Fig.~\ref{fig:architecture_metalearning}.
We train the backbone and FCR for multiple iterations, re-computing the class prototypes from meta-samples, $N$ randomly selected images per class.
After generating the class prototypes, we compute the cosine similarity ($\mathrm{cossim}$) between a query sample ($x$) and a class prototype ($\overline\theta_{p,i}$):
\begin{equation}\label{eq:meta_class_score}
    l_i = \mathrm{ReLU}(\mathrm{cossim}(FCR(f(x)), \overline\theta_{p,i}).
\end{equation}
Note that the backbone-extracted features are employed for $x$, whereas the prototypes, generated by clustering the features extracted using the same backbone, are stored in the~\gls{em}, as illustrated in Fig.~\ref{fig:architecture_metalearning}. 
Inspired by MANN~\cite{karunaratne2021mann}, we use the ReLU sharpening function to induce quasi-orthogonality.

Although~\gls{ce} shows good performance in previous studies~\cite{hersche2022constrained,song2023savcfscil}, it draws the features towards the cluster's centroid without accounting for the confidence level. 
This can lead to overfitting regions, where high confidence points are prioritized.
To this end, we employ a multi-margin loss defined as:
\begin{equation}
\label{eq:multi_margin_loss}
    \mathcal{L}_{MM}(l,y)={\frac{\sum_{i}\operatorname*{max}(0,(m-l_{gt}+l_i))^{2}}{|\mathcal{C}^0|}},
\end{equation}
where $y$ is the ground truth label and $m$ denotes margin value, which set to $m=0.1$ after grid search.
Multi-margin loss spreads the features near the classification frontier, improving the accuracy on distant points while maintaining its performance on those near the prototype. 

\begin{table*}[ht]
\centering
\caption{~\gls{ofscil} accuracy on CIFAR100. FP32 models were evaluated on NVIDIA GeForce GTX 1080 Ti, INT8 models were evaluated on GAP9~\gls{mcu}. FT represents optional iterative~\gls{fcr} fine-tuning. }
\label{tab:5_cifar100_comparison}
 
\begin{tabular}{lcp{20pt}cp{20pt}p{20pt}p{20pt}p{20pt}p{20pt}p{20pt}p{20pt}p{20pt}p{20pt}p{20pt}}
\toprule

 &   &  & \multicolumn{9}{c}{Session accuracy [\%]} &  \\
\cmidrule(r){5-13}

Method  &Backbone & Prec. & Size [MB] & \multicolumn{1}{c}{0} & \multicolumn{1}{c}{1} & \multicolumn{1}{c}{2} & \multicolumn{1}{c}{3} & \multicolumn{1}{c}{4} & \multicolumn{1}{c}{5} & \multicolumn{1}{c}{6} & \multicolumn{1}{c}{7} & \multicolumn{1}{c}{8} & \multicolumn{1}{c}{Avg.}                  \\ 
\cmidrule(r){1-4}\cmidrule(r){5-13}\cmidrule(r){14-14}

MetaFSCIL~\cite{chi2022metafscil}      & ResNet20 & FP32 & 1.08 & 74.50 & 70.10 & 66.84 & 62.77 & 59.48 & 56.52 & 54.36 & 52.56 & 49.97 & 60.79 \\
C-FSCIL~\cite{hersche2022constrained}  & ResNet12 & FP32 & 51.6 & 77.47 & 72.40 & 67.47 & 63.25 & 59.84 & 56.95 & 54.42 & 52.47 & 50.47 & 61.64 \\
LIMIT~\cite{zhou2022limit}             & ResNet20 & FP32 & 1.08 & 73.81 & 72.09 & 67.87 & 63.89 & 60.70 & 57.77 & 55.67 & 53.52 & 51.23 & 61.84 \\

SAVC~\cite{song2023savcfscil}          & ResNet12 & FP32 & 51.6 & 78.47 & 72.86 & 68.31 & 64.00 & 60.96 & 58.28 & 56.17 & 53.91 & 51.63 & 62.73 \\
ALICE~\cite{peng2022alice}             & ResNet18 & FP32 & 66.8 & 79.00 & 70.50 & 67.10 & 63.40 & 61.20 & 59.20 & 58.10 & 56.30 & 54.10 & 63.21 \\
NC-FSCIL~\cite{yang2023ncfscil}        & ResNet12 & FP32 & 51.6 & 82.52 & 76.82 & 73.34 & 69.68 & 66.19 & 62.85 & 60.96 & 59.02 & 56.11 & 67.50 \\

\cmidrule(r){1-4}\cmidrule(r){5-13}\cmidrule(r){14-14}

O-FSCIL           & \multirow{2}{*}{ResNet12} & FP32 & 51.6 & \textbf{84.05} & \textbf{79.10} & 74.23 & 69.96 & 66.92 & 63.89 & 61.67 & 59.51 & 57.10 & 68.52 \\

O-FSCIL + FT      &  & FP32 & 51.6 & 84.02 & 79.08 & \textbf{74.34} & \textbf{70.11} & \textbf{66.95} & \textbf{64.00} & \textbf{61.86} & \textbf{59.72} & \textbf{57.50} & \textbf{68.62} \\

\cmidrule(r){1-4}\cmidrule(r){5-13}\cmidrule(r){14-14}

\multirow{2}{*}{O-FSCIL}           & \multirow{2}{*}{MobileNetV2}  & FP32 & 10.0 & 77.51 & 72.82 & 68.41 & 64.25 & 61.24 & 57.98 & 55.32 & 53.03 & 50.73 & 62.37 \\
 & & INT8 & 2.5 & 76.97 & 72.46 & 68.24 & 64.19 & 61.07 & 58.14 & 56.01 & 53.55 & 51.37 & 62.44 \\
 \cmidrule(r){1-4}\cmidrule(r){5-13}\cmidrule(r){14-14}
\multirow{2}{*}{O-FSCIL}          & \multirow{2}{*}{MobileNetV2\_x2} & FP32 & 10.0 & 78.37 & 73.47 & 69.20 & 64.94 & 61.20 & 58.24 & 55.33 & 53.49 & 51.51 & 62.86 \\
 & & INT8 & 2.5 & 78.13 & 73.54 & 69.27 & 65.13 & 61.85 & 58.73 & 56.21 & 54.04 & 51.81 & 63.19 \\ 
\cmidrule(r){1-4}\cmidrule(r){5-13}\cmidrule(r){14-14}
\multirow{2}{*}{O-FSCIL}          & \multirow{2}{*}{MobileNetV2\_x4} & FP32 & 10.0 & 81.79 & 77.37 & 72.73 & 68.42 & 64.87 & 61.76 & 59.76 & 57.21 & 54.95 & 66.54 \\
& & INT8 & 2.5 & 81.56 & 76.81 & 72.56 & 68.32 & 64.95 & 61.94 & 59.65 & 57.47 & 55.33 & 66.51 \\
\cmidrule(r){1-4}\cmidrule(r){5-13}\cmidrule(r){14-14}
\multirow{2}{*}{O-FSCIL + FT}     & \multirow{2}{*}{MobileNetV2\_x4} & FP32 & 10.0 & 81.90 & 77.31 & 72.90 & 68.48 & 65.09 & 61.91 & 59.73 & 57.72 & 55.45 & 66.75 \\ 
& & INT8 & 2.5 & 81.53 & 76.82 & 72.54 & 68.44 & 64.81 & 61.76 & 59.65 & 57.78 & 55.72 & 66.56 \\
\bottomrule
\end{tabular}

\smallskip 

\vspace{-0.5cm}
    
\end{table*}

\section{Hardware deployment}
\label{sec:hardware}

\subsection{Deployment}
\label{ssc:deployment}

We deploy and evaluate our model on the GAP9~\gls{mcu}, a multi-core \gls{simd} processor suitable for vector and matrix computations, present in neural networks.
GAP9 comprises two processing components, a fabric controller, handling control and communications, and a 9-core cluster designed to efficiently execute parallelized algorithms.
All cores share access to an L1 memory and instruction cache, an L2 memory is shared by the compute processors, and the system also supports an external \SI{8}{\mega \byte} L3 memory.
Furthermore, \gls{dma} units enable asynchronous L1$\leftrightarrow$L2 and L2$\leftrightarrow$L3 memory transfers.

To deploy our model on GAP9, we first quantize the weights and activations of our metalearned~\gls{fp} model to 8-bit integers using~\gls{tqt} algorithm in Quantlib~\cite{spallanzani2022quantlab}, with additional quantization-aware pretraining and metalearning epochs.
The network is then deployed using Dory~\cite{burrello2020dory}, whereas an additional IO interface layer is implemented to support the evaluation of our~\gls{ofscil}.

\subsection{On-Device learning}
\label{ssc:ondevice}

During the deployment, the \gls{mcu} performs online novel class learning, computing the corresponding class prototype and storing it to~\gls{em}. 
For inference, a query image is classified according to the class prototype with the highest cosine similarity with the query $\theta_p$ feature. 
To increase the number of classes that can be learned on the device, we reduce the memory requirements of~\gls{em}.
We study precision reductions through bit-shift divisions in Section~\ref{ssc:deployment}.

Similar to Mode 2 in C-FSCIL~\cite{hersche2022constrained}, we implement an optional~\gls{fcr} finetuning, while freezing the backbone.
To minimize the training effort, we store average class activations $\overline\theta_{a,i}$ in an activation memory.
We then iteratively update the~\gls{fcr} by maximizing the similarity between the~\gls{fcr} mapping of $\overline\theta_{a,i}$ and the bipolarized class prototype through batched gradient descent over $B$ iterations. 
To minimize the memory accesses, we develop a sub-batching mechanism that creates three input matrices from $N$ pairs of $\overline\theta_{a,i}$, $\mathrm{FCR}(\overline\theta_{a,i})$, and $\overline\theta_{p,i}$.
This allows for computing the accumulated gradient of $N$ samples, reducing the number of memory accesses to $B/N$ per batch.

\section{Experimental results}
\label{sec:results}

\subsection{CIFAR100 benchmark}
\label{ssc:cifar100}

We use the CIFAR100~\cite{krizhevsky2009cifar100} dataset to evaluate our architecture, split into three sets: base session (50 images/class for 60 classes), class-incremental learning sessions (eight 5-way, 5-shot sessions) and 100 images per class for the test set.

As shown in Table~\ref{tab:5_cifar100_comparison},~\gls{ofscil} achieves state-of-the-art accuracy with ResNet-12 as a backbone, outperforming other works by more than one percentage point averaged over eight learning sessions.
Notably, our pretraining and metalearning lead to an accuracy gain of 1.5\% in the base session, paving the way for robust incremental learning in later stages.
This further enables us to employ~\gls{ofscil} on lightweight MobileNetv2 backbones, achieving average test accuracy levels of up to 66.54\% for MobileNetV2\_x4, outperformed only by 1\% by the larger~\gls{ncfscil}.
Notably, the minimal accuracy reduction comes with a 5.7$\times$ reduction in computational effort and a 5.2$\times$ decrease in storage requirements compared to the ResNet12 backbone, as shown in Table~\ref{tab:3_modified_mobilenetv2}.

Remarkably, our metalearning strategy allows us to generate robust, separable sample projections without expensive retraining of the~\gls{fcr}.
Nevertheless, if backpropagation-based fine-tuning is employed on MobileNetV2\_x4, an additional 0.2\% is to be gained, yet this would incur an adaptation cost of up to 6.6G~\glspl{mac}/session, compared to 0.7G~\glspl{mac}/session on MobileNetV2\_x4 without any fine-tuning. 
The computational effort reduction of 8.8$\times$ further motivates the extreme edge potential of our architecture.

\begin{table}[t]
\caption[Ablation studies on CIFAR100 benchmark]{Ablation study of the proposed methods for the accuracy[\%] on CIFAR100. AG: augmentation, OR: orthogonal regularization, MM: multi-margin-based metalearning, CE: cross-entropy-based metalearning, FT: incremental fine-tuning. The experiments were conducted with ResNet12 backbone.} \label{tab:5_ablation_study}
\begin{tabular}{>{\centering}p{0.7cm}>{\centering}p{0.7cm}>{\centering}p{0.7cm}>{\centering}p{0.7cm}>{\centering}p{0.7cm}|cc|c|}
\hline
\multirow{2}{*}{\textbf{AG}} & \multirow{2}{*}{\textbf{OR}} & \multirow{2}{*}{\textbf{MM}} & \multirow{2}{*}{\textbf{CE}} & \multirow{2}{*}{\textbf{FT}} & \multicolumn{2}{c|}{\textbf{Session}}        & \multirow{2}{*}{\textbf{Avg}} \\ \cline{6-7}
   &   &   &   &   & \multicolumn{1}{c}{\textbf{0}} & \textbf{8} &                               \\ \hline
   &   &   &   &   & \multicolumn{1}{c}{79.72}      & 51.47      & 62.94                         \\ \hline
\checkmark  &   &   &   &   & \multicolumn{1}{c}{81.87}      & 53.85      & 64.77                         \\ \hline
\checkmark  & \checkmark &   &   &   & \multicolumn{1}{c}{83.52}      & 56.72      & 67.88                         \\ \hline
\checkmark  &   & \checkmark &   &   & \multicolumn{1}{c}{83.65}      & 56.25      & 67.56                         \\ \hline
\checkmark  & \checkmark & \checkmark &   &   & \multicolumn{1}{c}{84.05}      & 57.10      & 68.52                         \\ \hline
\checkmark  & \checkmark &   & \checkmark &   & \multicolumn{1}{c}{83.02}      & 51.54      & 64.56                         \\ \hline
\checkmark  & \checkmark & \checkmark &   & \checkmark & \multicolumn{1}{c}{84.02}      & 57.50      & 68.62                         \\ \hline
\end{tabular}

\smallskip 

\vspace{-0.3cm}

\end{table}

\subsection{Ablation study}
\label{ssc:ablation}

This section investigates the benefits of our proposed pretraining and metalearning. 
As a baseline, we consider the ResNet-12-pretrained~\gls{ofscil} architecture.
Firstly, we notice that data augmentation generates accuracy gains of 2.15\% on the base session and 2.4\% on the eighth one.
Second, adding orthogonal regularization in the pipeline significantly boosts performance, particularly for the new classes, leading to accuracy increments between 1.65\% and 2.87\%.
This confirms that orthogonalization encourages neural networks to learn useful features beyond those of the base classes.
Interestingly,~\gls{ce} metalearning incurs performance degradation. 
Corroborating this with~\gls{ce} loss reductions noticed during training without the accuracy classification gains, we conclude that~\gls{ce} discourages feature robustness and generalization.

\subsection{Deployment}

\begin{table}[t]
    \centering
    \caption[The execution time, power, and energy of FSCIL modules]{The execution time, power, and energy consumption on GAP9 for~\gls{ofscil} --~\gls{em} update, emphasized -- and for~\gls{fcr} finetuning, added for comparison. The results are reported per class, for a five-shot learning setting. Finetuning is performed for 100 epochs. BB denotes the backbone. }\label{tab:5_forward_measurement}
    \begin{tabular}{|l|l|c|c|c|}
    \hline
    \multicolumn{1}{|c|}{\textbf{Operation}} & \multicolumn{1}{c|}{\textbf{BB}} & \multicolumn{1}{c|}{\textbf{Time [ms]}} & \multicolumn{1}{c|}{\textbf{Power [mW]}} & \multicolumn{1}{c|}{\textbf{Energy [mJ]}} \\ \hline
    FCR  &     $\forall$      & 3.23$^{\pm 0.73}$  & 47.75 $^{\pm0.34}$ & 0.15 $^{\pm 0.01}$ \\ \hline
    
    \multirow{3}{*}{BB inference} & M & 48.10$^{\pm 5.14}$ & 43.96 $^{\pm 0.98}$ & 2.12 $^{\pm 0.23}$ \\ 
    & M2 & 52.51 $^{\pm 5.27}$ & 45.12 $^{\pm0.24}$ & 2.40 $^{\pm 0.24}$ \\ 
    & M4   & 99.50 $^{\pm 2.41}$ & 44.19 $^{\pm 0.64}$ & 4.40 $^{\pm 0.12}$ \\ \hline
    \multirow{3}{*}{\textbf{EM update}} & M & 256.65$^{\pm11.6}$ & 44.22 $^{\pm 0.84}$ & 11.35 $^{\pm 0.24}$ \\
    & M2 & 278.70 $^{\pm 11.9}$& 45.75$^{\pm 0.26}$ & 12.75 $^{\pm 0.25}$ \\ 
    & M4 & 513.65 $^{\pm 5.63}$ & 44.29 $^{\pm 0.59}$ & 22.75 $^{\pm 0.12}$ \\ \hline
    \multirow{3}{*}{FCR finetune} & M & 6171.7 $^{\pm 29.8}$ & 50.29 $^{\pm 0.55}$ & 310.35$^{\pm 0.72}$ \\ 
    & M2 & 6193.7 $^{\pm 29.9}$ & 50.33 $^{\pm 0.52}$ & 311.75 $^{\pm 0.74}$\\ 
    & M4 & 6428.7 $^{\pm 28.0}$ & 50.05 $^{\pm 0.54}$ & 321.75 $^{\pm 0.58}$ \\ \hline

    \end{tabular} 

    \vspace{-0.5cm}

    \smallskip
    \justifying

\end{table}

We quantize~\gls{ofscil} using~\gls{tqt}~\cite{jain2020tqt}, with three pretraining epochs and ten metalearning iterations following the quantization.
As shown in Table~\ref{tab:5_cifar100_comparison}, similar accuracy levels are measured when comparing \texttt{int8}-quantized models with \texttt{fp32} networks.
Interestingly, quantized networks achieve higher accuracy for the latter sessions, as reducing the precision acts as a regularizer, improving the features' separability.

We furthermore measured the latency, power, and energy consumption for our quantized and deployed models, presented in Table~\ref{tab:5_forward_measurement}.
We deploy our models on the GAP9~\gls{mcu}, operating at~\SI{650}{\m \volt}, \SI{240}{\mega \hertz} as this is the most energy-efficient operating point for the \gls{mcu}.
Notably, we can perform both inference and training in real-time, as~\gls{ofscil} learns a new class only~\SI{256}{\m \s}.
Moreover, we remain within the~\SI{50}{\m \watt} power envelope also for backpropagation-based~\gls{fcr}.
By performing both ~\gls{em} update and last layer finetuning on MobileNetV2\_x4, GAP9 draws up to~\SI{320}{\m \joule} per new class.
Without the finetuning, it can efficiently learn new classes consuming ~\SI{12}{\m \joule} with minor degradation, demonstrating the feasibility of~\gls{ofscil} for battery-operated devices.

\definecolor{myblue}{RGB}{100,143,255}  
\definecolor{mypink}{RGB}{220,38,127} 
\definecolor{myyellow}{RGB}{255,176,0} 

\begin{figure}[!t]
\begin{tikzpicture}
    \begin{axis}[
        title=Backbone,
        width=0.6\linewidth,
        height=5cm,
        bar width= 4,
        ybar,
        ylabel={MACs/cycle},
        y label style={at={(axis description cs:0.2,.5)},anchor=south},
        legend style={legend columns=1},
        legend style={nodes={scale=0.9, transform shape}},
        legend pos=north west,
        ymin=0, ymax=7.5,
        ytick={0, 1, 2, 3, 4, 5, 6, 7},
        enlarge y limits =false,
        ymajorgrids = true,
        xtick=data,
        enlarge x limits= 0.2,
        xticklabels={1,2,4,8},
    ]
    \addplot[red,fill=red!30!white,error bars/.cd,y dir=both,y explicit,]
    coordinates {
      (1,0.7584835772579667) +- (0,0)
      (2,1.4026803909024355) +- (0,0)
      (3,2.1868996807127123) +- (0,0)
      (4,2.8854801497848250) +- (0,0)
    };
    \addplot[green,fill=green!30!white,error bars/.cd,y dir=both,y explicit,]
    coordinates {
      (1,0.9143187250112346) +- (0,0)
      (2,1.7558415002084784) +- (0,0)
      (3,3.1062005526757740) +- (0,0)
      (4,4.6235006828671334) +- (0,0)
    };
    \addplot[blue,fill=blue!30!white,error bars/.cd,y dir=both,y explicit,]
    coordinates {
      (1,1.108307341433482) +- (1.2876785019592952e-05,1.2876785019592952e-05)
      (2,2.157484874639854) +- (5.484911650265227e-05,5.484911650265227e-05)
      (3,3.8524095464759793) +- (0.0005959718610198847,0.0005959718610198847)
      (4,6.512480037831121) +- (0.0015518653800492778,0.0015518653800492778)
    };
    
    \addlegendentry{MobileNetV2}
    \addlegendentry{MobileNetV2\_x2}
    \addlegendentry{MobileNetV2\_x4}
\end{axis}
\end{tikzpicture}
\begin{tikzpicture}
    \begin{axis}[
        title=FCR,
        width=0.3\linewidth,
        height=5cm,
        bar width= 4,
        ybar,
        legend style={
        legend columns=2},
        ymin=0, ymax=0.75,
        ytick={0,0.1,0.2,0.3,0.4,0.5,0.6,0.7},
        enlarge y limits =false,
        ymajorgrids = true,
        xtick=data,
        enlarge x limits= 0,
        xticklabels={,1,2,4,8,},
    ]
    \addplot[purple,fill=purple!30!white,error bars/.cd,y dir=both,y explicit,]
    coordinates {
      (1,0) +- (0,0)
      (2,0.46237322573368717) +- (0.00018721592310870698,0.00018721592310870698)
      (3,0.46889574296177333) +- (0.00017880227639638428,0.00017880227639638428)
      (4,0.47044209610349360) +- (0.00017349967110392540,0.00017349967110392540)
      (5,0.46793019878988000) +- (0.00018206918083195170,0.00018206918083195170)
      (6,0) +- (0,0)
    };
\end{axis}
\end{tikzpicture}
\begin{tikzpicture}
    \begin{axis}[
        title=Finetune,
        width=0.3\linewidth,
        height=5cm,
        bar width= 4,
        ybar,
        legend style={
        legend columns=2},
        ymin=0, ymax=1.5,
        ytick={0, 0.2, 0.4, 0.6, 0.8, 1.0, 1.2, 1.4},
        enlarge y limits =false,
        ymajorgrids = true,
        xtick=data,
        enlarge x limits= 0,
        xticklabels={,1,2,4,8,},
    ]
    \addplot[purple,fill=purple!30!white,error bars/.cd,y dir=both,y explicit,]
    coordinates {
      (1,0) +- (0,0)
      (2,0.16679968788659727) +- (0.00027838128784387034,0.00027838128784387034)
      (3,0.32717764471443406) +- (0.00106209515635150320,0.00106209515635150320)
      (4,0.62557513288470750) +- (0.00392336536590835800,0.00392336536590835800)
      (5,1.08354635409302770) +- (0.01201307888728372700,0.01201307888728372700)
      (6,0) +- (0,0)
    };
\end{axis}
\end{tikzpicture}

\smallskip 
\centering

\caption[The average operation per cycle at different numbers of cores]{Average number of operations per cycle given the number of active cores, for backbone inference (left),~\gls{fcr} inference (centre), and~\gls{fcr} backpropagation update (right).}

\vspace{-0.3cm}

\label{fig:5_macs_operation}
\end{figure}
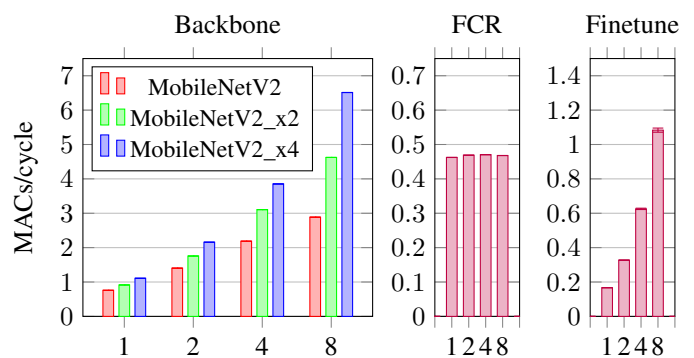

Fig.~\ref{fig:5_macs_operation} illustrates the impact of the multi-core architecture of GAP9 on~\gls{ofscil}. 
Highly parallelizable given the presence of convolutional layers, we can achieve up to \SI{6.5}{\glspl{mac}/cycle} for our largest MobileNetV2 backbone with 8 cores.
The parallelization potential reduces as we increase the number of strided convolutions in the feature extractor.
The performance gains of parallelizing the \gls{fcr} layer on a multi-core architecture are not as pronounced as those for other layers, primarily due to the data transfer overheads. 
Specifically, transferring approximately \SI{328}{\kilo\byte} of data from L3 to the L1 caches incurs a latency of roughly \SI{3}{}~ms. 
In contrast, when parallelized, the actual computation consumes only about \SI{0.25}{}~ms, showcasing a significant speedup of nearly 5 times on the multi-core system.

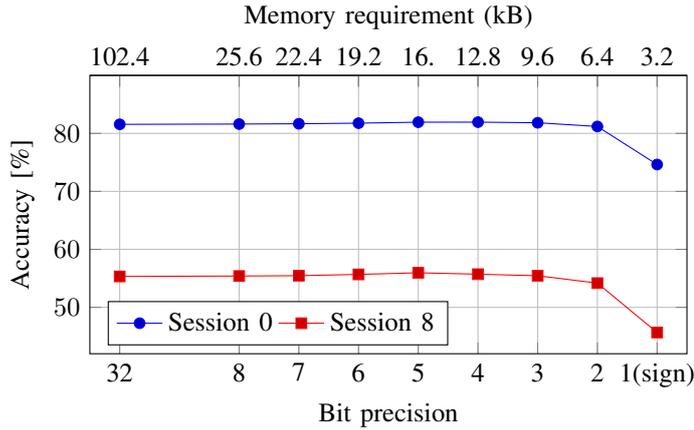
\begin{figure}[!t]
    \centering
    \begin{tikzpicture}
    \begin{axis}[
        xlabel={Bit precision},
        ylabel={Accuracy [\%]},
        y label style={at={(axis description cs:0.08,.5)},anchor=south},
        width=1.05\linewidth,
        height=150,
        legend pos = south west,
        ymin=42, ymax=90,
        xmin=-0.5, xmax=9.5,
        ytick={50,60,70,80},
        legend columns = 3,
        ymajorgrids=true,
        xmajorgrids=true,
        xticklabels={32, 8, 7, 6, 5, 4, 3, 2, 1(sign)},
        xtick={0,2,3,4,5,6,7,8,9},
    ]
    
    \addplot table [x=x, y=0, col sep=comma] {fig/5fig_EM_precision_tradeoff.dat};
    \addlegendentry{Session 0}
    \addplot table [x=x, y=8, col sep=comma] {fig/5fig_EM_precision_tradeoff.dat};
    \addlegendentry{Session 8}
    
    \end{axis}

    \begin{axis}[
        ymin=42, ymax=90,
        xmin=-0.5, xmax=9.5,
        axis x line*=top,
        axis y line=none,
        xlabel near ticks,
        xticklabels={102.4, 25.6,  22.4,  19.2,  16.,  12.8,   9.6,  6.4,  3.2},
        xtick={0,2,3,4,5,6,7,8,9},
        xlabel={Memory requirement (kB)},
        width=1.05\linewidth,
        height=150,
    ]
    \end{axis}
    
    \end{tikzpicture}
   
   \caption{The representation precision in the episodic memory impacts the accuracy of MobileNetV2\_x4-based model, considering 100 class prototypes stored in the memory.}
   \label{fig:5_EM_precision_tradeoff}

\vspace{-0.7cm}
   
\end{figure}

To reduce the memory requirements, we analyze the impact of the memory precision $\overline{\theta}_{p,i}$ on the accuracy, shown in Fig.~\ref{fig:5_EM_precision_tradeoff}.
A 17-bit integer is sufficient to represent the class prototype without overflow for MobileNetV2\_x4.
We can further reduce the $\overline{\theta}_{p,j}$ bit length to an 8-bit integer by performing a 9-bit right shift (i.e., vector division), reducing the norm while maintaining the general $\overline{\theta}_{p,j}$ vector direction and preserving the accuracy.
Further reductions down to 3-bit class prototypes can be achieved without accuracy drops, thus enabling us to store 100 class prototypes with only~\SI{9.6}{\kilo \byte}.

\section{Conclusion}
\label{sec:conclusion}

We proposed~\gls{ofscil}, a novel~\gls{fscil} methodology using orthogonal regularization and multi-margin-based metalearning to improve feature separability in incoming classes.
We achieved state-of-the-art accuracy on the CIFAR100 dataset using ResNet12 and comparable results with~\gls{ncfscil} when using the $5\times$ smaller and $3\times$ more computationally efficient MobileNetV2\_x4.  
We moreover designed, deployed, and evaluated ~\gls{ofscil} on GAP9~\gls{mcu}, with energy requirements to learn a new class as low as \SI{12}{\m \joule}, making it suitable for battery-operated extreme edge devices. 

\section{Acknowledgements}
\vspace{-0.1cm}
This work was partly supported by the Swiss National Science Foundation under grant No 207913: TinyTrainer: On-chip Training for TinyML devices.
\vspace{-0.1cm}
\bibliographystyle{IEEEtran}
\bibliography{main}

\end{document}